\documentclass{article}

\usepackage{times}
\usepackage{graphicx} 
\usepackage{subfigure}

\usepackage{natbib}

\usepackage{algorithm}
\usepackage{algorithmic}

\usepackage{hyperref}

\usepackage[accepted]{icml2017}

\icmltitlerunning{Cosine Normalization}

\begin{document}

\twocolumn[
\icmltitle{Cosine Normalization: Using Cosine Similarity Instead of Dot Product in Neural Networks}




\begin{icmlauthorlist}
\icmlauthor{Luo Chunjie}{ict,ucas}
\icmlauthor{Zhan Jianfeng}{ict}
\icmlauthor{Wang Lei}{ict}
\icmlauthor{Yang Qiang}{bafst}

\end{icmlauthorlist}

\icmlaffiliation{ict}{Institute of Computing Technology, Chinese Academy of Sciences}
\icmlaffiliation{ucas}{University of Chinese Academy of Sciences}
\icmlaffiliation{bafst}{Beijing Academy of Frontier Science and Technology}

\icmlcorrespondingauthor{Luo Chunjie}{luochunjie@ict.ac.cn}
\icmlcorrespondingauthor{Zhan Jianfeng}{zhanjianfeng@ict.ac.cn}
\icmlcorrespondingauthor{Wang Lei}{wanglei\_2011@ict.ac.cn}
\icmlcorrespondingauthor{Yang Qiang}{yangqiang@mail.bafst.com}

\icmlkeywords{cosine normalization, neural networks}

\vskip 0.3in
]



\printAffiliationsAndNotice{}  

\begin{abstract}
Traditionally, multi-layer neural networks use  dot product between the output vector of previous layer and the incoming weight vector as the input to activation function.
The result of dot product is unbounded, thus increases the risk of large variance.
Large variance of neuron makes the model sensitive to the change of input distribution, thus results in poor generalization, and aggravates the internal covariate shift which slows down the training.
To bound dot product and decrease the variance, we propose to use cosine similarity or centered cosine similarity (Pearson Correlation Coefficient) instead of dot product in neural networks, which we call cosine normalization.
We compare cosine normalization with batch, weight and layer normalization in fully-connected neural networks as well as convolutional networks on the data sets of MNIST, 20NEWS GROUP, CIFAR-10/100 and SVHN.
Experiments show that cosine normalization achieves better performance than other normalization techniques.
\end{abstract}

Deep neural networks have received great success in recent years in many areas, e.g. image recognition \cite{krizhevsky2012imagenet}, speech processing \cite{hinton2012deep}, natural language processing \cite{mikolov2013distributed}, Go game \cite{silver2016mastering}.
Training deep neural networks is nontrivial task. Gradient descent is commonly used to train neural networks.
However, due to gradient vanishing problem \cite{hochreiter2001gradient}, it works badly when directly applying to deep networks.

Lots of approaches have been adopted to overcome the difficulty of training deep networks. For example, pre-training \cite{hinton2006fast,hinton2006reducing}, special network structure \cite{simonyan2014very,szegedy2015going,he2016deep}, ReLU activation \cite{nair2010rectified,maas2013rectifier},  noise injecting  \cite{wan2013regularization,srivastava2014dropout}, normalization \cite{ioffe2015batch,salimans2016weight,ba2016layer,arpit2016normalization,ren2016normalizing}.

In previous work, multi-layer neural networks use dot product (also called inner product) between the output vector of previous layer and the incoming weight vector as the input to activation function.
\begin{equation} \label{dot product}
net=\vec{w} \cdot \vec{x}
\end{equation}
where $net$ is the input to activation function (pre-activation),  $\vec{w}$ is the incoming weight vector, and $\vec{x}$ is the input vector which is also the output vector of previous layer, ($\cdot$) indicates dot product.
Equation \ref{dot product} can be rewritten as Equation \ref{dot product2}, where $\cos \theta$ is the cosine of angle between $\vec{w}$ and $\vec{x}$, $\left|~\right|$ is the Euclidean norm of vector.
\begin{equation} \label{dot product2}
net=\left|\vec{w}\right|  \left|\vec{x}\right| \cos \theta
\end{equation}

The result of dot product is unbounded, thus increases the risk of large variance.
Large variance of neuron makes the model sensitive to the change of input distribution, thus results in poor generalization.
Large variance could also aggravate the internal covariate shift  which slows down the training \cite{ioffe2015batch}.
Using small weights can alleviate this problem. Weight decay (L2-norm) \cite{krogh1991simple} and max normalization (max-norm) \cite{srebro2005rank,srivastava2014dropout} are methods that could decrease the weights.
Batch normalization \cite{ioffe2015batch} uses statistics calculated  from mini-batch training examples to normalize the result of dot product, while layer normalization \cite{ba2016layer} uses statistics from the same layer on a single training case. The variance can be constrained within certain range using batch or layer normalization.
 Weight normalization \cite{salimans2016weight} re-parameterizes the weight vector by dividing its norm,  thus partially bounds the result of dot product.

To thoroughly bound dot product, a straight-forward idea is to use cosine similarity. Similarity (or distance) based methods are widely used in data mining and machine learning \cite{tan2006introduction}.
Particularly, cosine similarity is most commonly used in high dimensional spaces. For example, in information retrieval and text mining, cosine similarity gives a useful measure of how similar two documents are \cite{singhal2001modern}.

In this paper, we combine cosine similarity with neural networks. We use cosine similarity instead of dot product when computing the pre-activation. That can be seen as a normalization procedure, which we call cosine normalization. Equation \ref{cosine normalization} shows the cosine normalization.
\begin{equation} \label{cosine normalization}
net_{norm}= \cos \theta = \frac{\vec{w} \cdot \vec{x}} {\left|\vec{w}\right|  \left|\vec{x}\right|}
\end{equation}
To extend, we can use the centered cosine similarity, Pearson Correlation Coefficient (PCC), instead of dot product.
By ignoring the magnitude of $\vec{w}$ and $\vec{x}$,  the input to activation function is bounded between -1 and 1. Higher learning rate could be used for training without the risk of large variance.
Moreover, network with cosine normalization can be trained by both batch gradient descent and stochastic gradient descent, since it does not depend on any statistics on batch or mini-batch examples.

We compare our cosine normalization with batch, weight and layer normalization in fully-connected neural networks on the MNIST and 20NEWS GROUP data sets. Additionally, convolutional networks with different normalization techniques are evaluated on the  CIFAR-10/100 and SVHN data sets. Here is a brief summary:
\begin{itemize}
\item Cosine normalization achieves lower test error than batch, weight and layer normalization
\item Centered cosine normalization ( Pearson Correlation Coefficient ) further reduces the test error.
\item Cosine normalization is more stable than other normalization techniques, specially batch normalization.
\item Cosine normalization can accelerate neural networks training as well as other normalization.
\end{itemize}

\section{Background and Motivation}

Large variance of neuron in neural network makes the model sensitive to the change of input distribution, thus results in poor generalization. Moreover, variance could be amplified as information moves forward along layers, especially in deep network.  Large variance could also aggravate the internal covariate shift, which refers the change of distribution of each layer during training, as the parameters of previous layers change \cite{ioffe2015batch}. Internal covariate shift slows down the training  because the layers need to continuously adapt to the new distribution.
Traditionally, neural networks use dot product to compute the pre-activation of neuron. The result of dot product is unbounded. That is to say, the result could be any value in the whole real space,  thus increases the risk of large variance.

Using small weights can alleviate this problem, since the pre-activation $net$ in Equation \ref{dot product2}  will be decreased when $\left|\vec{w}\right|$ is small. Weight decay \cite{krogh1991simple}  and max normalization \cite{srebro2005rank,srivastava2014dropout} are methods that try to make the weights to be small. Weight decay adds an extra term to the cost function that penalizes the squared value of each weight separately. Max normalization puts a constraint on the maximum squared length of the incoming weight vector of each neuron. If update violates this constraint, max normalization scales down the vector of incoming weights to the allowed length.
The objective (or direction to objective) of original optimization problem is changed when using weight decay (or max normalization). Moreover, they bring additional hyper parameters that should be carefully preset.

Batch normalization \cite{ioffe2015batch} uses statistics calculated from mini-batch training examples to normalize the pre-activation. The normalized value is re-scaled and re-shifted using additional parameters.
Since batch normalization uses the statistics on mini-batch examples, its effect is dependent on the mini-batch size.
To overcome this problem, normalization propagation \cite{arpit2016normalization} uses a data-independent parametric estimate of mean and standard deviation,
while layer normalization \cite{ba2016layer} computes the mean and standard deviation from the same layer on a single training case.
Weight normalization \cite{salimans2016weight} re-parameterizes the incoming weight vector by dividing its norm. It decouples the length of weight vector from its direction, thus partially bounds the result of dot product. But it does not consider the length of input vector. These methods all bring additional parameters to be learned, thus make the model more complex.

An important source of inspiration for our work is cosine similarity, which is widely used in data mining and machine learning \cite{singhal2001modern,tan2006introduction}.
To thoroughly bound dot product, a straight-forward idea is to use cosine similarity. We combine cosine similarity with neural network, and the details will be described in the next section.

\section{Cosine Normalization}

\begin{figure}[!htb]
\centering
\includegraphics[width=0.4\columnwidth]{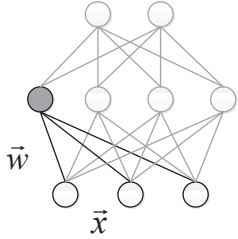}
\caption{A simple neural network. The output of hidden unit is the nonlinear transform of dot product between input vector  and incoming weight vector. That is computed by $f(\vec{w}\cdot\vec{x})$. With cosine normalization, The output of hidden unit is computed by $f(\frac{\vec{w} \cdot \vec{x}} {\left|\vec{w}\right|  \left|\vec{x}\right|})$ }
\label{neural}
\end{figure}

To decrease the variance of neuron, we propose a new method, called cosine normalization, which simply uses cosine similarity instead of dot product in neural network.
A simple multi-layer neural network is shown in Figure \ref{neural}.
Using cosine normalization, the output of hidden unit  is computed by Equation \ref{neuron_out}.
\begin{equation} \label{neuron_out}
o = f(net_{norm})= f(\cos \theta) = f(\frac{\vec{w} \cdot \vec{x}} {\left|\vec{w}\right|  \left|\vec{x}\right|})
\end{equation}
where $net_{norm}$ is the normalized pre-activation,  $\vec{w}$ is the incoming weight vector and $\vec{x}$ is the input vector, ($\cdot$) indicates dot product, $f$ is nonlinear activation function.
Cosine normalization bounds the pre-activation between -1 and 1. The result could be even smaller when the dimension is high. As a result, the variance can be controlled within a very narrow range.

Empirically, we find that using ReLU activation function $max(0,net_{norm})$, the result of normalization needs no re-scaling and re-shifting. Therefore, there is no additional parameter to be learned or hyper-parameter to be preset. However, when using other activation functions , like sigmoid, tanh, or softmax, the result of normalization should be re-valued to fully utilize the non-linear regime of the functions.

When implementing of cosine normalization in fully-connected nets, we just need divide the norm of incoming weight vector, as well as the norm of input vector. The input vector is the output vector of previous layer. That is to say, the hidden units in the same layer have the same norm of input vector. While in the convolutional nets, the input vector is constrained in a receptive field. Different receptive fields have different norms.

One thing should be noticed is that cosine similarity can only measure the similarity between two non-zero vectors, since denominator can not be zero. Non-zero bias can be added to avoid the situation of zero vector. Let $\vec{w}=[w_{1},w_{2}...w_{i}]$ and $\vec{x}=[x_{1},x_{2}...x_{i}]$. After adding bias, $\vec{w}=[w_{0},w_{1},w_{2}...w_{i}]$ and $\vec{x}=[x_{0},x_{1},x_{2}...x_{i}]$,  where $w_{0}$ and $x_{0}$ should be non-zero.

We can use gradient descent (back propagation) to train the neural network with cosine normalization.
Comparing to batch normalization, cosine normalization does not depend on any statistics on batch or mini-batch examples, so the model can be trained by both batch gradient descent and stochastic gradient descent. Meanwhile, cosine normalization performs the same computation in forward propagation at training and inference times.
The procedure of back propagation in neural network with cosine normalization is the same as ordinary neural network except the derivative of $net_{norm}$ with respect to $w$ or $x$.

To show the derivative conveniently, dot product can be rewritten as Equation \ref{dot_product3},  where $w_{i}$ indicates the i dimension of vector $\vec{w}$, and $x_{i}$ indicates the i dimension of vector $\vec{x}$.
\begin{equation} \label{dot_product3}
net= \sum_{i}(w_{i}x_{i})
\end{equation}
Therefore, the derivative of $net$ with respect to $w_{i}$ or $x_{i}$ in ordinary neural network can be calculated by Equation \ref{diff_w_dot} or Equation \ref{diff_x_dot}.
\begin{eqnarray} \label{diff_w_dot}
\frac{\partial net}{\partial w_{i}} = x_{i}
\end{eqnarray}
\begin{eqnarray}  \label{diff_x_dot}
\frac{\partial net}{\partial x_{i}} = w_{i}
\end{eqnarray}

Correspondingly, the cosine normalization can be rewritten as Equation \ref{cosine normalization2}.
\begin{equation} \label{cosine normalization2}
net_{norm}= \cos \theta = \frac{\sum_{i}(w_{i}x_{i})} {\sqrt{\sum_{i}(w_{i}^2)}\sqrt{\sum_{i}(x_{i}^2)}}
\end{equation}
Then,  the derivative of $net_{norm}$ with respect to $w_{i}$ or $x_{i}$  can be calculated by Equation \ref{diff_w1} or Equation \ref{diff_x1}.
\begin{eqnarray} \label{diff_w1}
\frac{\partial net_{norm}}{\partial w_{i}} =\frac{x_i}{\sqrt{\sum_{i}(w_{i}^2)}\sqrt{\sum_{i}(x_{i}^2)}} - \nonumber \\ \frac{w_{i}\sum_{i}(w_{i}x_{i})}{(\sqrt{\sum_{i}(w_{i}^2)})^3\sqrt{\sum_{i}(x_{i}^2)}}
\end{eqnarray}
\begin{eqnarray} \label{diff_x1}
\frac{\partial net_{norm}}{\partial x_{i}} =\frac{w_i}{\sqrt{\sum_{i}(w_{i}^2)}\sqrt{\sum_{i}(x_{i}^2)}} - \nonumber \\ \frac{x_{i}\sum_{i}(w_{i}x_{i})}{\sqrt{\sum_{i}(w_{i}^2)}(\sqrt{\sum_{i}(x_{i}^2)})^3}
\end{eqnarray}
Equation \ref{diff_w1} or Equation \ref{diff_x1} can be briefly written as Equation \ref{diff_w2} or Equation \ref{diff_x2}.
\begin{equation} \label{diff_w2}
\frac{\partial net_{norm}}{\partial w_{i}} =\frac{x_i}{\left|\vec{w}\right| \left|\vec{x}\right|} - \frac{w_{i}(\vec{w} \cdot \vec {x})}{\left|\vec{w}\right|^3 \left|\vec{x}\right| }
\end{equation}
\begin{equation} \label{diff_x2}
\frac{\partial net_{norm}}{\partial x_{i}} =\frac{w_{i}}{\left|\vec{w}\right| \left|\vec{x}\right|} - \frac{x_i(\vec{w} \cdot \vec {x})}{\left|\vec{w}\right| \left|\vec{x}\right|^3 }
\end{equation}

As pointed in \cite{lecun2012efficient}, centering the inputs of units can help the training of neural networks.
Batch or layer normalization centers the data by subtracting the mean of batch or layer,
while mean-only batch normalization can enhance the performance of weight normalization \cite{salimans2016weight}.
We can use Pearson Correlation Coefficient (PCC), which is centered cosine similarity, to extend cosine normalization:
\begin{equation} \label{Pearson correlation}
net_{norm}=  \frac{(\vec{w}-\mu_{w}) \cdot (\vec{x}-\mu_{x})} {\left|\vec{w}-\mu_{w}\right|  \left|\vec{x}-\mu_{x}\right|}
\end{equation}
where $\mu_{w}$ is the mean of $\vec{w}$ and $\mu_{x}$ is the mean of  $\vec{x}$.

\section{Discussions}

\subsection{Comparing to weight normalization}
Weight normalization \cite{salimans2016weight}  re-parameterizes the weights by using new parameters as:
\begin{equation} \label{weight normalization}
\vec{w}_{new}=  \frac{g}{\left|\vec{w}\right|} {\vec{w}}
\end{equation}
Then, the output of hidden unit  is computed as:
\begin{equation} \label{weight_out}
o = f(net_{norm})= f( \vec{w}_{new} \cdot \vec{x}) = f \left(\frac{g}{\left|\vec{w}\right|} {\vec{w}} \cdot \vec{x} \right)
\end{equation}
where $g$ is a re-scaling parameter and can be learned by gradient descent.
Ignoring the re-scaling parameter $g$, weight normalization could be seen as partial cosine normalization which only constrains the weights. By additionally dividing the magnitude of  $\vec{x}$,  cosine normalization bounds pre-activation within a narrower range, thus makes lower variance of neurons.

Moreover, cosine normalization makes the model more robust for different input magnitude. For example, in the forward procedure of  the fully-connected network, we have $ \vec{x_{l+1}}= f(\vec{w} \cdot \vec{x_{l}})$. If we scale the $\vec{x_{l}}$ by a factor $\lambda$,
then $ \vec{x_{l+1}}= f(\vec{w} \cdot (\lambda \vec{x_{l}}))$. When the activation function f is ReLU,  we have $ \vec{x_{l+1}}= \lambda f(\vec{w} \cdot  \vec{x_{l}})$. So the $\lambda$ is linearly transmitted to the last layer. When the last layer is softmax,  $\exp(\vec{x})/\sum{\exp(\vec{x}})$, the output distribution becomes more steep due to the nonlinearity of softmax. For example, if the input vector to softmax is [1, 2], then the output distribution is [0.2689, 0.7311]. When the $\lambda=10$, after the linearly transmitting, the input vector to softmax becomes [10, 20], and the output distribution becomes [0, 1]. Supposing we want to recognize a handwritten digit, scaling the whole digit by a factor does not bring any valid information. In other words, the output distribution should not be changed. By using cosine normalization, the output distribution can be stable when the input magnitude varies,  and it depends only on the angle between the input and the weight.

In the backward procedure of weight normalization, the derivative of $net_{norm}$ with respect to $w_{i}$ or $x_{i}$  can be calculated by Equation \ref{diff_w1_wn} or Equation \ref{diff_x1_wn}.
\begin{eqnarray} \label{diff_w1_wn}
\frac{\partial net_{norm}}{\partial w_{i}} =\frac{x_i}{\left|\vec{w}\right|} - \frac{w_{i}(\vec{w} \cdot \vec {x})}{\left|\vec{w}\right|^3}
\end{eqnarray}
\begin{eqnarray} \label{diff_x1_wn}
\frac{\partial net_{norm}}{\partial x_{i}} =\frac{w_i}{\left|\vec{w}\right|}
\end{eqnarray}
After scaling the input by $\lambda$, the derivative of $net_{norm}$ with respect to $w_{i}$ becomes Equation \ref{diff_w1_wn_2}. Comparing  Equation \ref{diff_w1_wn} with Equation \ref{diff_w1_wn_2},  we can see that the scaling of input also makes the gradient scaling in weight normalization. While in cosine normalization, as shown in
Equation \ref{diff_w2}, the scaling factor $\lambda$ can be offset by the $\left|\lambda \vec{x}\right|$ in the denominator.
\begin{eqnarray} \label{diff_w1_wn_2}
{\frac{\partial net_{norm}}{\partial w_{i}}} =\frac{\lambda x_i}{\left|\vec{w}\right|} - \frac{w_{i}(\vec{w} \cdot \lambda \vec { x})}{\left|\vec{w}\right|^3} \nonumber \\
= \lambda \left( \frac{x_i}{\left|\vec{w}\right|} - \frac{w_{i}(\vec{w} \cdot \vec {x})}{\left|\vec{w}\right|^3} \right)
\end{eqnarray}

\subsection{Comparing to layer normalization}
Layer normalization \cite{ba2016layer} use Equation \ref{batch_out} to normalize pre-activation, followed by re-scaling and re-shifting the normalized value (Equation \ref{batch_out2}).
\begin{equation} \label{batch_out}
 net_{norm}== \frac{net-\mu}{\sigma} = \frac{\vec{w} \cdot \vec{x} - \mu}{\sigma}
\end{equation}
\begin{equation} \label{batch_out2}
o = f(\gamma ~ net_{norm} + \beta)
\end{equation}
The mean $\mu$ and standard deviation $\sigma$  are computed over a layer on a single training case. The $\gamma$ is re-scaling parameter and $\beta$ is re-shifting parameter, which are learned during training.

Because $\left|\vec{x}-\mu_{x}\right|= \sqrt{\sum_{i}(x_{i}-\mu_{x})^2}$, and $\sigma_{x}= \sqrt{\frac{1}{n}\sum_{i}(x_{i}-\mu_{x})^2}$, where $n$ is a constant referring to the dimension of $\vec{x}$,
The centered cosine normalization ( Pearson Correlation Coefficient ) can be re-write as:
\begin{equation} \label{pcc}
net_{norm}=  \frac{(\vec{w}-\mu_{w}) \cdot (\vec{x}-\mu_{x})} {n\sigma_{w}\sigma_{x}}
\end{equation}
Ignoring the constraining of weights, layer normalization is similar with Pearson Correlation Coefficient by constraining the $\vec{x}$ in fully-connected networks.

However,  there are three differences between Pearson Correlation Coefficient and layer normalization: 1) Pearson Correlation Coefficient constrains $\vec{w}$ as well as $\vec{x}$, while layer normalization constrains only $\vec{x}$. Thus Pearson Correlation Coefficient is robust to the scaling or shifting of both weight and input. 2) Layer normalization computes the mean and standard deviation before activation and after dot product, while Pearson Correlation Coefficient computes the mean and standard deviation before dot product and after activation. 3) In convolutional networks, Pearson Correlation Coefficient calculates the mean and standard deviation from the receptive fields, while layer normalization calculates the mean and standard deviation from the whole layer. That is to say, different receptive fields have different mean and standard deviation using Pearson Correlation Coefficient, while the same layer has the same  mean and standard deviation using layer normalization. As pointed in \cite{ba2016layer},  layer normalization works well when all the hidden units in a layer make similar contributions, while the assumption of similar contributions is no longer true for convolutional networks. Pearson Correlation Coefficient just needs the assumption of similar contributions in the receptive fields rather than the whole layer. That is more reasonable for the convolutional networks.

\subsection{Similarity metric in neural networks}
In  machine learning and data mining, there are lots of metrics to measure the similarity or distance between different samples. Among them, cosine similar or the centered cosine (Pearson Correlation Coefficient), is heavily used in many fields, e.g.  K-nearest neighbors for classification, K-means for clustering, information retrieval, item or user based recommendation. There are also some neural networks using similarity metrics as the output of neurons, e.g. Radial Basis Function networks (RBF) \cite{moody1989fast}, Self-Organizing Map (SOM) \cite{kohonen1982self}. The training of these networks is not using back propagation, and it is hard to build end-to-end deep networks using RBF or SOM. The paper \cite{lin2013network} argues that the level of abstraction is low with dot product (generalized linear model), thus uses multi-layer perceptron (network in network) to learn convolution filter in convolutional networks. Since dot product is not a decent metric, we may directly try other metrics.
As far as we know, it is the first time to use cosine similarity or Pearson Correlation Coefficient as the basic metric to build end-to-end deep network trained by back propagation.

\section{Experiments}
We compare our cosine normalization and centered cosine normalization (PCC) with batch, weight and layer normalization in fully-connected neural networks on the MNIST and 20NEWS GROUP data sets. Additionally, convolutional networks with different normalization are evaluated on the  CIFAR-10, CIFAR-100 and SVHN data sets. We also test the networks without any normalization both for fully-connected and convolutional.  The results are much worse than with normalization, thus we focus only on comparison of different normalization techniques.

\subsection{Date sets}

\subsubsection{MNIST}
The MNIST \cite{lecun1998gradient} data set consists of 28x28 pixel  handwritten digit black and white images. The task is to classify the images into 10 digit classes.
There are  60, 000 training  images and 10, 000 test images in the MNIST data set.
We scale the pixel values to the [0, 1] range  before inputting to our models.

\subsubsection{20NEWS GROUP}
The original training set contains 11269 text documents, and the test set contains 7505 text documents.  Each document is classified into one topic out of 20. For convenience of using mini-batch gradient descent, 69 examples in training set and 5 examples in test set are randomly dropped. As a result, there are 11200 training examples and 7500 test examples in our experiments. The words whose document frequency is larger than 5  are used as the input features. There are 21567 feature dimensions finally. Then,  the model of Term Frequency-Inverse Document Frequency (TF-IDF) is used to transform the text documents into vectors. After that, each feature is re-scaled to the range of [0, 1].

\subsubsection{CIFAR-10/100}
CIFAR-10 \cite{krizhevsky2009learning} is a data set of natural 32x32 RGB images
in 10-classes with 50, 000 images for training and 10, 000 for testing. CIFAR-100 is similar with CIFAR-10 but with 100 classes.
To augment data, the images are cropped to 24 x 24 pixels,  centrally for evaluation or randomly for training.
Then, a series of random distortions are applied:
1) randomly flip the image from left to right. 2) randomly distort the image brightness. 3) randomly distort the image contrast. The procedure of augmentation is the same as CIFAR-10 example in Tensorflow \cite{abadi2016tensorflow}.

\subsubsection{SVHN}
The Street View House Numbers (SVHN) \cite{netzer2011reading} dataset includes
604, 388 images (both training set and extra set) and 26, 032 testing images. Similar
to MNIST, the goal is to classify the digit centered in each 32x32 RGB image. We augment the data using the same procedure as CIFAR-10/100 mentioned above.

\subsection{Protocols}
A fully-connected neural network which has two hidden layers is used in experiments of MNIST and 20NEWS GROUP. Each hidden layer has 1000 units. The last layer is the softmax classification layer with 10-class for MNIST,  and 20-class for 20NEWS GROUP. To evaluated the convolutional networks, as shown in Table \ref{arch}, a VGG-like architecture with 12 weighted layers is evaluated in experiments of CIFAR-10/100 and SVHN. Each convolutional layer has 3$\times$3 receptive field with a stride of 1, and each max pool layer has 2$\times$2 regions with a stride of 1.
\begin{table}[!htb]
\caption{ VGG-like architecture.}
\label{arch}
\centering
\begin{tabular}{c}
\hline
conv-512 \\
conv-512 \\
conv-512 \\
maxpool \\
conv-512 \\
conv-512 \\
conv-512 \\
maxpool \\
conv-512 \\
conv-512 \\
conv-512 \\
maxpool  \\
fully-connected-1000 \\
fully-connected-1000 \\
fully-connected-10/100 \\
soft-max \\
\hline
\end{tabular}
\end{table}

ReLU activation function is used in the hidden layers. All weights are randomly initialized by truncated normal distribution with 0 mean and 0.1 variance. Mini-batch gradient descent is used to train the networks. The batch size is 100 in experiments of fully-connected nets, and 128 in convolutional nets. In our experiments, we use no re-scaling and re-shifting after normalization for hidden layers. However, for the last layer, we re-scale the normalized values before inputting to softmax. We tried different learning rate for all normalization techniques, and found that cosine normalization can use larger learning rate than other normalization techniques. The learning rate of the cosine normalization, centered cosine normalization (PCC), batch normalization, weight normalization, layer normalization is 10, 10, 1, 1, 1, respectively in our experiments. The exponential moving average of parameters with 0.9999 is used during inference in convolutional networks. No any regularization, dropout, or dynamic learning rate is used. We train the fully-connected nets with 200 epochs and convolutional nets $10^5$ step since the performances are not improved anymore (in this paper, training epoch refers a cycle that all training data are used once for training, while training step refers the times of update for parameters).

\subsection{Results}

\subsubsection{MNIST}
\begin{figure}[!htb]
\centering
\includegraphics[width=0.8\columnwidth]{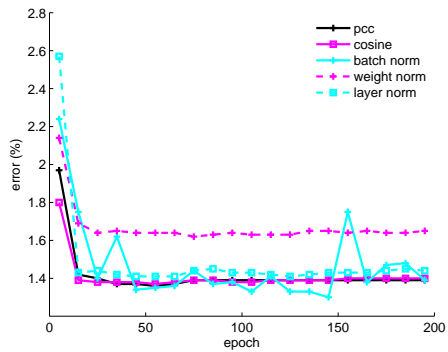}
\caption{The MNIST test error of different normalization techniques, vs. the number of training epoch. }
\label{mnist_fig}
\end{figure}

The results of test error for MNIST are shown in Figure \ref{mnist_fig}. As we can see, the converging speeds for different normalization techniques are close. That observation is also true for other data sets we will present next. That is to say, cosine normalization can accelerate the  training of networks as well as other normalization. We can also observe that centered cosine normalization (Pearson Correlation Coefficient) and cosine normalization achieve similar test errors, and which are slightly better than layer normalization. Table \ref{mnist_table} shows the mean and variance of test error for the last 50 epochs. Centered cosine normalization achieves the lowest mean of test error 1.39\%, while cosine and layer normalization achieve 1.40\%, 1.43\%   respectively. Weight normalization has the highest test error comparing to other normalization. Although batch normalization gets lowest test error at some point, it causes large variance of test error as training continues. Large fluctuation of batch normalization is caused by the change of statistics on different mini-batch examples.

\begin{table}[!htb]
\caption{ The mean and variance of test error in last 50 epochs in MNIST experiments. }
\label{mnist_table}
\centering
\begin{tabular}{lcc}
\hline
methods & mean \% & variance ($10^{-3}$) \\
\hline
centered cosine (PCC)    & 1.39 & 0  \\
cosine norm & 1.40 & 0.009 \\
batch norm    & 1.45 & 6.740   \\
weight norm    & 1.65  &   0.054      \\
layer norm     &  1.43 & 0.108 \\
\hline
\end{tabular}
\end{table}

\subsubsection{20NEWS GROUP}

\begin{figure}[!htb]
\centering
\includegraphics[width=0.8\columnwidth]{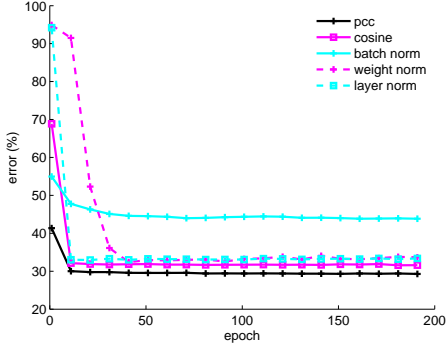}
\caption{The 20NEWS GROUP test error of  different normalization techniques, vs. the number of training epoch.}
\label{20news_fig}
\end{figure}

The results for 20NEWS GROUP are shown in Figure  \ref{20news_fig} and Table \ref{20news_table}.
Centered cosine normalization achieves the lowest test error 29.37\%, and cosine normalization achieves the second lowest test error 31.73\%. The batch normalization performs poorly in this task of high dimensional text classification. It only achieves 43.94\% test error.  Weight normalization (33.55\%) and layer normalization (33.29\%) achieve close performances. Both batch and weight normalization have larger variance of test error than other normalization.

\begin{table}[!htb]
\caption{ The mean and variance of test error in last 50 epochs in 20NEWS experiments. }
\label{20news_table}
\centering
\begin{tabular}{lcc}
\hline
methods & mean \% & variance ($10^{-2}$)  \\
\hline
centered cosine (PCC)    & 29.37 & 0.201  \\
cosine norm & 31.73 & 0.633 \\
batch norm    & 43.94  & 1.231  \\
weight norm    & 33.55  &   4.775      \\
layer norm     &  33.29 & 0.556 \\
\hline
\end{tabular}
\end{table}

\subsubsection{CIFAR-10}

\begin{figure}[!htb]
\centering
\includegraphics[width=0.8\columnwidth]{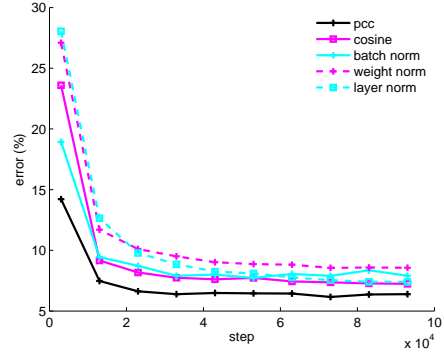}
\caption{The CIFAR-10 test error of  different normalization techniques, vs. the number of training step.}
\label{cifar10_fig}
\end{figure}

The results for CIFAR-10 are shown in Figure  \ref{cifar10_fig} and Table \ref{cifar10_table}.
Centered cosine normalization achieves the lowest test error 6.39\%, and cosine normalization achieves the second lowest test error 7.33\%. The layer normalization also achieves good performance, better than batch normalization,  in this experiment. It achieves 7.42\% test error.  Batch normalization achieves test error 8.08\%, and  still has larger variance of test error than other normalization. Weight normalization achieves the highest test error 8.55\%.

\begin{table}[!htb]
\caption{ The mean and variance of test error in last 10000 step in CIFAR-10 experiments. }
\label{cifar10_table}
\centering
\begin{tabular}{lcc}
\hline
methods & mean \% & variance ($10^{-3}$) \\
\hline
centered cosine (PCC)    & 6.39 & 0.076  \\
cosine norm & 7.33 & 0.036 \\
batch norm    & 8.08 & 1.052  \\
weight norm    & 8.55  &  0.010      \\
layer norm     &  7.42 & 0.008 \\
\hline
\end{tabular}
\end{table}

\subsubsection{CIFAR-100}

\begin{figure}[!htb]
\centering
\includegraphics[width=0.8\columnwidth]{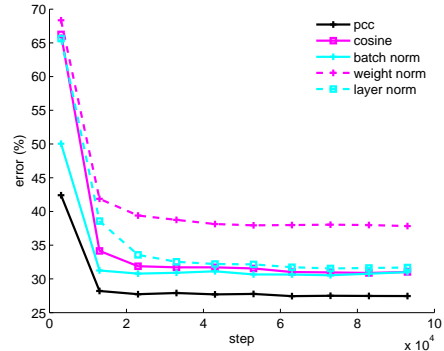}
\caption{The CIFAR-100 test error of  different normalization techniques, vs. the number of training step.}
\label{cifar100_fig}
\end{figure}

The results for CIFAR-100 are shown in Figure  \ref{cifar100_fig} and Table \ref{cifar100_table}.
Centered cosine normalization achieves the lowest test error 27.49\%. Cosine normalization and batch normalization achieve very close performance, 31.02\% and 31.01\% respectively. But batch normalization have  larger variance of test error. Weight normalization achieves the highest test error 37.87\%.

\begin{table}[!htb]
\caption{ The mean and variance of test error in last 10000 step in CIFAR-100 experiments. }
\label{cifar100_table}
\centering
\begin{tabular}{lcc}
\hline
methods & mean \% & variance ($10^{-4}$) \\
\hline
centered cosine (PCC)    & 27.49  & 1.03  \\
cosine norm & 31.02 & 0.43 \\
batch norm    & 31.01  & 3.23  \\
weight norm    & 37.87  &   1.36      \\
layer norm     &  31.66 & 0.22 \\
\hline
\end{tabular}
\end{table}

\subsubsection{SVHN}

\begin{figure}[!htb]
\centering
\includegraphics[width=0.8\columnwidth]{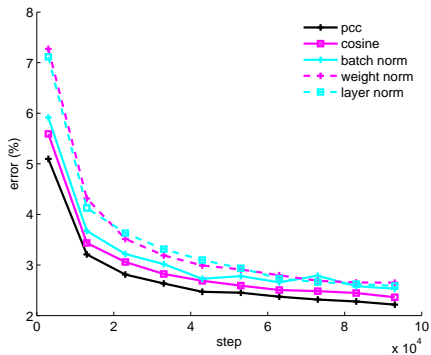}
\caption{The SVHN test error of  different normalization techniques, vs. the number of training step.}
\label{svhn_fig}
\end{figure}

The results for SVHN are shown in Figure  \ref{svhn_fig} and Table \ref{svhn_table}.
Centered cosine normalization achieves the lowest test error 2.22\%, and cosine normalization achieves the second lowest test error 2.34\%. Batch and layer normalization achieve test error 2.49\%, 2.58\% respectively. Weight normalization has the highest test error 2.63\%.

\begin{table}[!htb]
\caption{ The mean and variance of test error in last 10000 step in SVHN experiments. }
\label{svhn_table}
\centering
\begin{tabular}{lcc}
\hline
methods & mean \% & variance ($10^{-4}$)  \\
\hline
centered cosine (PCC)    & 2.22 & 0.01  \\
cosine norm & 2.34  & 0.11 \\
batch norm    & 2.49 & 0.14  \\
weight norm    & 2.63  & 0.03      \\
layer norm     &  2.58 & 0.01 \\
\hline
\end{tabular}
\end{table}

\section{Conclusions}

In this paper, we propose a new normalization technique, called cosine normalization, which uses cosine similarity or centered cosine similarity, Pearson correlation coefficient, instead of dot product in neural networks. Cosine normalization bounds the pre-activation of neuron  within a narrower range, thus makes lower variance of neurons. Moreover, cosine normalization makes the model more robust for different input magnitude.
Networks with cosine normalization can be trained using back propagation. It does not depend on any statistics on batch or mini-batch examples, and performs the same computation in forward propagation at training and inference times. In convolutional networks, it normalizes the neurons from the receptive fields rather than the same layer or batch size. Cosine normalization is evaluated on different types of network (fully-connected network and convolutional network) and on different data sets (MNIST, 20NEWS GROUP, CIFAR-10/100, SVHN). Experiments show that cosine normalization and centered cosine normalization significantly reduce the test error of classification comparing to batch, weight and layer normalization.



\bibliography{lcj_paper}
\bibliographystyle{icml2017}

\end{document}